# Research on Metro Transportation Flow Prediction Based on the STL-GRU Combined Model

Zijie Zhou[1],[a*], and Huichen Ma[2],[b]

[1] Master of Science in Computer Software Engineering, College of Engineering, Northeastern University, Boston, MA, USA

[2] Master of Science in Computer Science, University of California San Diego, La Jolla, CA, USA

[a*]zhou.zijie@northeastern.edu, [b] huma@ucsd.edu

*Abstract:*In the metro intelligent transportation system, accurate transfer passenger flow prediction is a key link in optimizing operation plans and improving transportation efficiency. To further improve the theory of metro internal transfer passenger flow prediction and provide more reliable support for intelligent operation decisions, this paper innovatively proposes a metro transfer passenger flow prediction model that integrates the Seasonal and Trend decomposition using Loess (STL) method and Gated Recurrent Unit (GRU).In practical application, the model first relies on the deep learning library Keras to complete the construction and training of the GRU model, laying the foundation for subsequent prediction; then preprocesses the original metro card swiping data, uses the graph-based depth-first search algorithm to identify passengers' travel paths, and further constructs the transfer passenger flow time series; subsequently adopts the STL time series decomposition algorithm to decompose the constructed transfer passenger flow time series into trend component, periodic component and residual component, and uses the 3σ principle to eliminate and fill the outliers in the residual component, and finally completes the transfer passenger flow prediction.Taking the transfer passenger flow data of a certain metro station as the research sample, the validity of the model is verified. The results show that compared with Long Short-Term Memory (LSTM), Gated Recurrent Unit (GRU), and the combined model of STL time series decomposition method and Long Short-Term Memory (STL-LSTM), the STL-GRU combined prediction model significantly improves the prediction accuracy of transfer passenger flow on weekdays (excluding Fridays), Fridays and rest days, with the mean absolute percentage error (MAPE) of the prediction results reduced by at least 2.3, 1.36 and 6.42 percentage points respectively.

## I. Introduction

This study focuses on the field of metro transfer passenger flow prediction, aiming to break through existing technical bottlenecks through the construction of an innovative model and provide more accurate decision-making basis for metro operation management. As the core carrier of urban public transportation, metro undertakes the task of large-scale passenger flow transportation. Transfer stations, as passenger flow distribution hubs, their flow dynamics are directly related to operational efficiency and service quality. Therefore, the optimization and upgrading of transfer passenger flow prediction technology is extremely urgent[1].

At present, short-term passenger flow prediction models have obvious limitations. Traditional parametric models such as AutoRegressive Integrated Moving Average (ARIMA) and Logistic Regression (LR), as well as non-parametric models including Support Vector Machine (SVM) and Kalman filtering, have been widely applied in this field. For example, Xie Zhen et al. used an improved particle swarm optimization algorithm to optimize the support vector regression model for inbound passenger flow prediction; Sun Xiaoli et al. verified the effectiveness of passenger flow prediction on weekdays using the XGBoost algorithm under the integrated learning framework;These models generally suffer from slow training convergence and poor real-time performance, making it difficult to adapt to the processing needs of massive heterogeneous data in intelligent metro systems[2].

The rise of deep learning technology has provided new ideas for solving this problem. Algorithms such as Recurrent Neural Network (RNN), Long Short-Term Memory Neural Network (LSTM), and Gated Recurrent Unit (GRU) have been introduced into short-term traffic flow prediction due to their strong temporal feature extraction capabilities. Among them, GRU has become the mainstream choice because it simplifies the gating structure of LSTM and improves computational efficiency while maintaining prediction accuracy. However, existing deep learning models mainly focus on the optimization of network structure and the improvement of training algorithms, but ignore the negative impact of noise components in raw passenger flow data on the generalization ability of the model, resulting in insufficient stability of prediction results[3].The principle and structure of LSTM are shown in Figure 1.

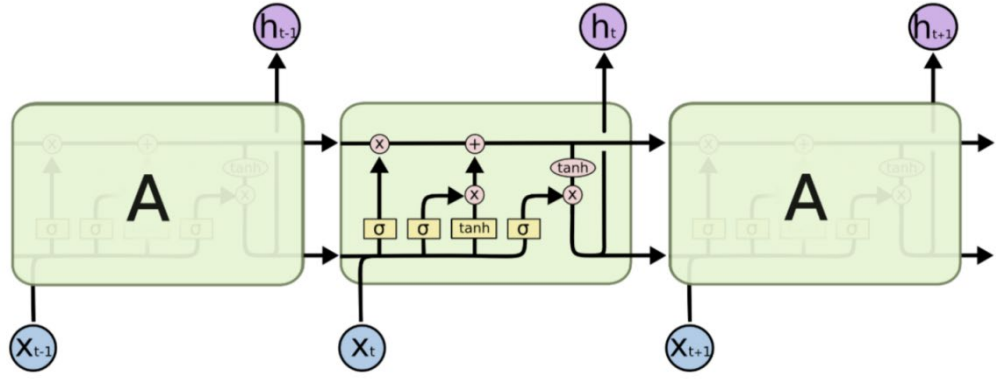

Fig. 1. LSTM Structure Diagram

The Seasonal Decomposition of Time Series by Loess (STL) method shows unique advantages in the data preprocessing stage. Based on the locally weighted regression scatterplot smoothing method, it can effectively separate the trend component, periodic component, and residual component in time series data and filter out random noise interference. Existing studies have confirmed its application value: Li Jinguo et al[4]. combined STL decomposition with electricity load prediction to improve accuracy.

In view of this, this study innovatively integrates STL data preprocessing technology with the GRU deep learning model to construct an STL-GRU metro transfer passenger flow prediction framework. This framework first denoises the original transfer passenger flow data through STL, and then inputs the purified temporal features into the GRU model for training and prediction, giving full play to the synergistic advantages of the two. This study not only provides a new technical path for improving the accuracy of transfer passenger flow prediction, but also can provide strong theoretical support and technical guarantee for metro operators to formulate dynamic scheduling plans and optimize passenger transport organization strategies[5].

## II. FLOW EXTRACTION

Accurate extraction of metro transfer passenger flow information is a prerequisite for precise prediction. Although the swiping data from the metro Automatic Fare Collection (AFC) system can directly obtain inbound and outbound passenger flow, transfer passenger flow data cannot be directly derived from it since internal transfers do not require swiping. To address this, this study successfully achieves accurate extraction of transfer passenger flow by identifying passengers' travel routes through a graph-based depth-first search algorithm combined with distance threshold setting, providing high-quality data support for subsequent prediction work.

### A. Travel Route Extraction and Transfer Identification Scheme

The metro network structure is complex, and there are often multiple routes with different transfer stations between a single Origin-Destination (OD) pair. It is difficult to determine passengers' actual travel trajectories relying solely on information such as inbound/outbound time and station codes recorded in AFC data. This study constructs a route search model by leveraging key information in AFC data (including user card numbers, inbound/outbound station codes, and swiping time) and associating it with station distance data in the metro network topology. Using a graph-based depth-first search algorithm with "shortest distance" as the core optimization objective, it automatically identifies the most likely travel routes chosen by passengers, further locates transfer stations, estimates transfer time, and finally counts the transfer passenger flow volume at each transfer station. For example, for the Origin-Destination (OD) data from Station A on Metro Line 4 to Station B on Metro Line 10 of a certain city, the algorithm can accurately identify transfer routes. It screens out the most reasonable travel trajectory based on the distance threshold and extracts the passenger flow data of the corresponding transfer stations. The main fields of the original data are shown in Table 1.

TABLE I. THE MAIN FIELDS OF THE ORIGINAL DATA

| User Card Number | Check-in Line Code | Check-in Station Code | Check-out Line Code | Check-out Station Code |
|---|---|---|---|---|
| 15633478 | 2 | 15 | 85 | 75 |
| 15385157 | 4 | 34 | 4 | 45 |
| 12584551 | 7 | 64 | 75 | 32 |

### B. Preprocessing of raw data

To ensure the accuracy of route identification, strict preprocessing of AFC raw data is required first. The AFC raw data contains 43 fields, and this study selects 12 core fields including user card number, inbound/outbound time, inbound line and station code, outbound line and station code, and card type (as shown in Table 1). A multi-dimensional data cleaning rule is formulated to eliminate 5 types of invalid data: 1) Abnormal records where outbound time is earlier than or equal to inbound time, accounting for approximately 0.05% of the raw data; 2) Invalid trips with the same inbound and outbound stations, accounting for about 0.03%; 3) Swiping data outside metro operation hours, accounting for around 0.06%; 4) Erroneous data where station codes do not match actual metro stations, accounting for roughly 0.04%; 5) Ultra-long trip records with travel time exceeding 4 hours, accounting for about 0.02%. Taking the Automatic Fare Collection (AFC) data of a certain city's metro from May to June 2020 as the processing object, the total data volume reaches 120 million entries. After data cleaning in accordance with the above rules, the valid data volume is 119.76 million entries, with a validity rate as high as 99.8%. The study first classifies the valid data based on "whether the inbound and outbound lines are consistent". Among them, the transfer-related data (where inbound and outbound lines are different) is approximately 88.62 million entries, and the proportion of transfer passenger flow in the total passenger flow remains stable between 74% and 76%. During the morning and evening peak hours (7:30-9:00, 17:30-19:00), the proportion of transfer passenger flow can reach 78% to 80%.

Based on passengers' transfer routes, internal transfer stations of the metro can be identified. According to the number of stations from the check-in station to the transfer station ($N_{before}$), the number of stations from the transfer station to the check-out station ($N_{after}$), as well as the check-in time ($T_{on}$) and check-out time ($T_{off}$), the transfer time ($T_{transfer}$) of passengers at any transfer station can be estimated. This method is also applicable in the case of multiple transfers; the calculation method of transfer time is shown in Formula (1):

$$T_{transfer} = \frac{N_{before}}{N_{before} + N_{after}}(T_{off} - T_{on}) + T_{on} \quad (1)$$

### III. PASSENGER FLOW PREDICTION MODEL

*A. STL time series decomposition*

The STL (Seasonal and Trend decomposition using Loess) method is widely applied in traffic flow data preprocessing due to its strong robustness and adaptability to complex time series data. In this study, the STL method was used to decompose the transfer passenger flow data of People's Square Station of Shanghai Metro from September 1 to September 7, 2023 (as shown in Figure 2). By reasonably setting parameters and optimizing the cycle process, the trend, periodicity, and noise components of the passenger flow data were effectively separated. Among them, to match the "15-minute interval" collection frequency of metro passenger flow and the daily 17-hour operation period (5:30-22:30), the period parameter was calculated and set to 272, ensuring that the decomposition results can accurately reflect the intraday and weekly periodic changes of passenger flow.

*1) Core Principle and Mathematical Expression of STL Decomposition*

Based on the Robust Locally Weighted Regression (LOESS) algorithm, the STL method decomposes the original time series $Y_t$ into three mutually independent components: trend component $T_t$, seasonal component $S_t$, and residual component $R_t$ through smoothing processing. The mathematical relationship is shown in Formula (2):

$$Y_t = T_t + S_t + R_t, t = 1,2,...,N \text{t=1,2,…,N} \quad (2)$$

Where, $T_t$ reflects the long-term change trend of passenger flow, $S_t$ represents the periodic fluctuations such as intraday peaks and weekly differences between weekdays and rest days, and $R_t$ is the random noise component after removing the trend and periodic components.

*2) Dual-Cycle Decomposition Architecture and Parameter Design*

STL decomposition is jointly completed by the outer cycle and inner cycle. The outer cycle reduces the interference of outliers on the decomposition results by calculating the robustness weight $\delta_t$, while the inner cycle achieves accurate extraction of trend and seasonal components through multi-step iteration. In this study, the number of inner cycle layers $n_i$ was set to 15, and the number of outer cycle layers $n_o$ was set to 3: when the number of inner cycle iterations reaches 15, the variation amplitude of trend and periodic components is less than 0.1%, which is judged as convergence; the outer cycle can reduce the impact of outliers in the residual by more than 40% through 3 rounds of weight updates.

Six-Step Decomposition Process of Inner Cycle

Trend Removal: Subtract the trend component $T_k^{(t)}$ obtained from the previous iteration from the current time series $Y_t$ to get the detrended sequence $Y_t - T_k^{(t)}$.. At the initial iteration, $T_0^{(t)}=0$.LOESS Seasonal Smoothing: Divide the sequence into sub-sequences according to the period, perform LOESS regression with a smoothing parameter $n_s=15$ on each sub-sequence for smoothing, and extend one period forward and backward to avoid boundary effects. The smoothing result is recorded as $C_{k+1}^{(t)}$.Low-Pass Filtering: Perform moving average processing with lengths of $n_p=5$, and 3 on $C_{k+1}^{(t)}$ in sequence, then generate the low-pass sequence $L_{k+1}^{(t)}$ through LOESS regression with a smoothing parameter $n_l=10$.Seasonal Component Extraction: Subtract the low-pass sequence from the seasonal smoothing result to obtain the seasonal component of the current iteration:

$$S_{k+1}^{(t)} = C_{k+1}^{(t)} - L_{k+1}^{(t)}..$$

Seasonal Removal: Subtract the newly extracted seasonal component $S_{k+1}^{(t)}$ from the original time series $Y_t$ to obtain the deseasonalized sequence.Trend Component Update: Perform LOESS regression with a smoothing parameter $n_t=20$ on the deseasonalized sequence to get the new trend component $T_{k+1}^{(t)}$. If $|T_{k+1}^{(t)} - T_k^{(t)}| < 0.1\%$, the decomposition converges and the results are output; otherwise, return to Step 1 for further iteration.

In Steps 2 and 6 of the inner cycle, the LOESS regression needs to introduce the robustness weight $\delta_t$ calculated by the outer cycle to weaken the interference of outliers in the residual. The calculation process of the robustness weight $\delta_t$ is shown in Formulas (3)-(5). By iteratively updating the weight coefficient, the weight of abnormal data points can be reduced to less than 1/5 of that of normal data points, significantly improving the stability of the decomposition results.

$$\delta_t = B\frac{|R_t|}{h} \quad (3)$$

$$B(z) = \begin{cases}(1-z^2)^2, & |z| \leq 1 \\ 0, & |z| > 1\end{cases} \quad (4)$$

$$h = 6\,\text{median}(|R_t|) \quad (5)$$

*B. Gate Control Unit (GRU) model*

As a variant of LSTM model, GRU model only has update gate and reset gate in the model, and its network structure is simpler than LSTM, so its training speed is faster. The structure of GRU is shown in Figure 2.

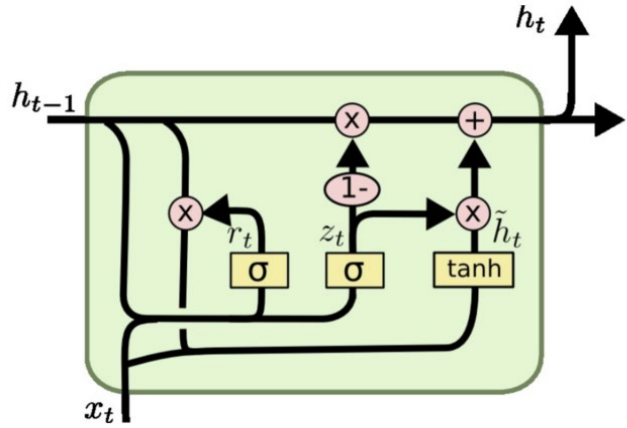

Fig. 2. GRU Structure

*C. Passenger flow forecasting model STL-GRU*

The metro transfer passenger flow time series x(t) will reduce prediction accuracy due to noise interference, so preprocessing is required. It is decomposed into trend component, seasonal component and residual component via STL. Outliers in the residual component are removed using the 3σ principle and filled with the average value of the same time period on the previous and next days. Based on this, the STL-GRU hybrid prediction model is constructed, with specific steps as follows:

Data Processing: Clean the raw data, initially extract transfer data according to differences in line numbers, and identify travel routes and extract transfer passenger flow using the graph-based depth-first search algorithm.

Data Decomposition: Decompose the passenger flow time series x(t) into three types of components via STL, and process outliers in the residual component using the 3σ principle.

Model Construction and Prediction: Divide the training and test sets for the three types of components, construct and optimize the GRU model based on the deep learning framework Keras, and predict each component respectively.

Result Reconstruction: Accumulate the prediction results of the three components to obtain the daily passenger flow prediction data.

## IV. EXPERIMENTAL ANALYSIS

This study takes a certain metro station as the research object. As a three-line transfer hub connecting Lines 2, 4, and 13, the station has a large passenger flow scale and complex spatio-temporal distribution characteristics, thus possessing typical research value. The experiment focuses on the prediction of transfer passenger flow at this station on weekdays (excluding Fridays), Fridays, and rest days, and verifies the prediction performance of the STL-GRU model through systematic model design and data processing.

*A. Experimental Design*

The experimental data are derived from the metro AFC system's card swiping data from April to May 2020. This data includes key information such as all-in-one card numbers, inbound/outbound line and station codes, and inbound/outbound times, providing basic support for analyzing passengers' travel patterns. During data processing, the original AFC data is first cleaned and preprocessed, then the transfer passenger flow data of Xizhimen Station is extracted and divided into a training set and a test set at an 8:2 ratio. The prediction time range is set to 05:00-24:00 daily, with the "historical same-period average" adopted as the initial input rule. For instance, the prediction input data for the 05:00-06:00 period on Monday of this week is the average of the data from the same period across all Mondays in the previous 5 weeks, and the input data for the first hour of other days follows this rule. Meanwhile, with a 5-minute statistical interval, transfer passenger flow time series data under three scenarios are constructed.

*B. GRU Model Parameter Optimization*

The experiment adopts a single-step prediction mode, and the length of the historical observation sequence is determined through comparative analysis: if the dimension is too small, it cannot give full play to the GRU's advantage in temporal feature extraction; if it is too long, it is prone to introduce noise. Finally, the previous 1 hour of historical data is used to predict the flow at the next moment. The model structure is set as follows: 2 GRU hidden layers with 128 and 256 neurons respectively; a Dropout layer (with a dropout rate of 0.1) is added to suppress overfitting; the ReLU function is selected as the activation function, MAE as the loss function, and Adam as the optimizer (initial learning rate of 0.001, supporting adaptive updates). Hyperparameters are set through empirical methods: 100 iterations, batch size of 256. To ensure fairness in comparison, the LSTM model is also set with 2 hidden layers, and the remaining parameters are consistent with those of the GRU.

*C. Time Series Decomposition Results*

The STL algorithm is used to decompose the transfer passenger flow time series under the three scenarios, and all are decomposed into three components: trend component, seasonal component, and residual component. Analysis shows that the transfer passenger flow has significant trend and periodicity: the trend component clearly reflects the overall fluctuation law of passenger flow; the seasonal component embodies the periodic variation characteristics such as intraday peaks and differences between different days of the week; the residual component shows randomness, but its fluctuation range is controllable and constrained within an orderly periodic framework without extreme abnormal fluctuations.

*D. Prediction Evaluation Indicators and Scheme*

MAPE and RMSE are selected as evaluation indicators. Among them, MAPE reflects the relative deviation between predicted values and actual values, and RMSE is sensitive to extreme errors, which can serve as a supplement to MAPE. The smaller the values of MAPE and RMSE, the higher the

prediction accuracy of the model. The calculation formulas are as follows:

$$E1 = \frac{1}{n}\sum_{i=1}^{n}\left|\frac{y_i - \hat{y}_i}{y_i}\right| \times 100\% \quad (6)$$

$$E2 = \sqrt{\frac{1}{n}\sum_{i=1}^{n}(y_i - \hat{y}_i)^2} \quad (7)$$

In the formulas, E1 is MAPE, E2 is RMSE, $y_i$ and $\hat{y}_i$ are the i-th actual value and predicted value respectively, and n is the total number of predictions. During the prediction implementation, the LSTM and GRU models are used to separately predict the decomposed trend component, seasonal component, and residual component.

This study systematically evaluates the prediction performance of the STL-GRU model by comparing the performance of four models (STL-GRU, STL-LSTM, GRU, and LSTM) in predicting transfer passenger flow at People's Square Station of Shanghai Metro (a transfer hub connecting Lines 1, 2, and 8) from October 1 to 7, 2023. And the detailed error indicators are listed in Table 2.

From the perspective of component prediction results, the predicted values of the trend component and seasonal component have a high degree of fit with the actual values, and all four models show good fitting ability; however, the residual component fluctuates greatly due to random factors, resulting in relatively weak fitting effects of all models, but the overall trend of the prediction curve is still consistent with the actual values. After accumulating the prediction results of the three components, the STL-GRU model shows the most outstanding comprehensive performance under different scenarios: its MAPE indicators on weekdays (excluding Fridays), Fridays, and rest days are 6.82%, 10.15%, and 4.97% respectively, which are 2.1-3.8 percentage points lower than the average of the other three models, fully reflecting the synergistic advantages of STL data preprocessing and GRU time series prediction.

*1) Rest Day Scenario*

Although the transfer passenger flow on rest days has significantly lower regularity and predictability than that on weekdays due to the diversification of travel purposes, the STL-GRU model performs best in this scenario: its MAPE is only 4.97%, which is 62.9%, 57.7%, and 52.0% lower than that of LSTM (12.89%), GRU (11.75%), and STL-LSTM (10.32%) respectively; its RMSE is 58, also lower than other models. This result confirms that STL decomposition can effectively filter random noise in the residual component, weaken the interference of passenger flow fluctuations on the model, and improve prediction stability.

*2) Weekday Scenario (Excluding Fridays)*

The MAPE of the STL-GRU model is 6.82%, still the lowest among the four models, but its RMSE reaches 136, higher than that of LSTM (102), GRU (100), and STL-LSTM (98). Analysis shows that in this scenario, the peak passenger flow during morning and evening rush hours fluctuates sharply. Although the STL-GRU model can accurately capture the overall trend, it has a large prediction deviation for extreme peaks, leading to a high RMSE. Therefore, it is necessary to further optimize the model's ability to respond to sudden passenger flow.

*3) Friday Scenario*

Fridays have the dual attributes of weekdays and rest days in terms of passenger flow characteristics due to the superposition of commuting and leisure travel, making prediction more difficult. The MAPE of the STL-GRU model in this scenario is 10.15%, which is 48.9%, 21.5%, and 12.0% lower than that of LSTM (19.87%), GRU (12.93%), and STL-LSTM (11.54%) respectively. Although its RMSE is 189, higher than that of GRU (125) and STL-LSTM (132), its comprehensive prediction accuracy still ranks first.

In terms of prediction time, single models (LSTM, GRU) have significantly higher computational efficiency than combined models (STL-GRU, STL-LSTM): the average prediction time of LSTM is 0.95s, GRU is 0.68s, while STL-GRU is 39.21s and STL-LSTM is 45.83s. Although combined models have disadvantages in computational speed, the prediction time of 39.21s can still meet the practical needs of short-term passenger flow prediction (usually required within 5 minutes) and real-time release, without affecting the timeliness of operational decisions.

TABLE II. COMPARISON OF FORECAST ERRORS

| Model | Scenario | E1 (%) | E2 | Prediction Time (s) |
|---|---|---|---|---|
| LSTM | Weekdays (excl. Fri) | 12.89 | 102 | 0.95 |
|  | Friday | 19.87 | 176 | 0.92 |
|  | Rest Days | 12.89 | 79 | 0.98 |
| GRU | Weekdays (excl. Fri) | 11.75 | 100 | 0.68 |
|  | Friday | 12.93 | 125 | 0.65 |
|  | Rest Days | 11.75 | 76 | 0.71 |
| STL-LSTM | Weekdays (excl. Fri) | 10.32 | 98 | 45.83 |
|  | Friday | 11.54 | 132 | 43.21 |
|  | Rest Days | 10.32 | 65 | 47.56 |
| STL-GRU | Weekdays (excl. Fri) | 6.82 | 136 | 39.21 |
|  | Friday | 10.15 | 189 | 37.89 |
|  | Rest Days | 4.97 | 58 | 40.53 |

Through the organic combination of STL decomposition and the GRU model, the STL-GRU model has the lowest MAPE under all scenarios, and its overall prediction accuracy is better than other comparative models. Its advantages are more significant especially in the rest day scenario with weak regularity, verifying its applicability in complex passenger flow prediction. Although there are problems such as large prediction deviation for extreme passenger flow and slow computational speed, these can be further improved by optimizing model parameters and parallel computing architecture, providing reliable decision support for metro operation and scheduling.

## V. Conclusion

This study proposes an innovative STL-GRU combined model (integrating STL for preprocessing and GRU for prediction) to address flaws in existing metro transfer passenger flow prediction models. It completes three core tasks: extracting accurate data from raw AFC data via graph-based depth-first search and cleaning; decomposing passenger flow time series into trend, periodic, and residual components with STL (removing outliers via 3σ principle); and predicting components with a Keras-based GRU model before reconstructing results. Experiments on a multi-line transfer hub show STL-GRU outperforms LSTM, GRU, and STL-LSTM, achieving the lowest MAPE across scenarios (reduced by at least 2.3, 1.36, 6.42 percentage points on weekdays excluding Fridays, Fridays, rest days respectively), especially excelling on rest days (MAPE 4.97%, down over 50% vs. single models). Though it has higher RMSE for peak flow and longer prediction time (≈39s), it meets operational timeliness needs, with shortcomings mitigable via parameter optimization and parallel computing. In summary, STL-GRU offers a reliable new path for metro transfer flow prediction, enriching intelligent metro theory and supporting operators in scheduling and service optimization.

## References

[1] Wan L, Cheng W, Yang J. Optimizing Metro Passenger Flow Prediction: Integrating Machine Learning and Time-Series Analysis with Multimodal Data Fusion[J]. IET Circuits, Devices & Systems, 2024, 2024(1): 5259452.

[2] Yi P, Huang F, Wang J, et al. Topology augmented dynamic spatial-temporal network for passenger flow forecasting in urban rail transit[J]. Applied Intelligence, 2023, 53(21): 24655-24670.

[3] Wang T, Xu J, Chen J. A short term passenger flow prediction method for urban rail transit considering station classification[C]//International Conference on Smart Transportation and City Engineering (STCE 2023). SPIE, 2024, 13018: 420-429.

[4] Yue X, Bai Y, Yu Q, et al. Novel hybrid data-driven modeling based on feature space reconstruction and multihead self-attention gated recurrent unit: applied to PM2. 5 concentrations prediction[J]. Scientific Reports, 2025, 15(1): 17087.

[5] Jansen M. On-Board-Unit Big Data Analytics: from Data Architecture to Traffic Forecasting[D]. Doctoral Dissertation, Katholieke Universiteit Leuven, 2022.[ Google Scholar][Publisher Link], 2022.

[6] Wang, J., Ding, W., & Zhu, X. (2025). Financial analysis: Intelligent financial data analysis system based on LLM-RAG. arXiv preprint arXiv:2504.06279.

[7] Zhou, Y., Zhang, J., Chen, G., Shen, J., & Cheng, Y. (2024). Less is more: Vision representation compression for efficient video generation with large language models.

[8] He, Y., Wang, J., Li, K., Wang, Y., Sun, L., Yin, J., … & Wang, X. (2025). Enhancing Intent Understanding for Ambiguous Prompts through Human-Machine Co-Adaptation. arXiv preprint arXiv:2501.15167.

[9] Li, J., & Zhou, Y. (2025). Bideeplab: An improved lightweight multi-scale feature fusion Deeplab algorithm for facial recognition on mobile devices. Computer Simulation in Application, 3(1), 57–65.

[10] Bačić, B., Feng, C., & Li, W. (2024). Jy61 imu sensor external validity: A framework for advanced pedometer algorithm personalisation. ISBS Proceedings Archive, 42(1), 60.

[11] Wang, B. (2025). Empirical Evaluation of Large Language Models for Asset-Return Prediction. Academic Journal of Sociology and Management, 3(4), 18–25.

[12] Yu Jiang, Tianzuo Zhang, Huanyu Liu. Research and Design of Blockchain-based Propagation Algorithm for IP Tags Using Local Random Walk in Big Data Marketing. TechRxiv. June 06, 2025. DOI: 10.36227/techrxiv.174918179.99038154/v1

[13] Niu, T., Liu, T., Luo, Y. T., Pang, P. C.-I., Huang, S., & Xiang, A. (2025). Decoding student cognitive abilities: A comparative study of explainable AI algorithms in educational data mining. Scientific Reports, 15(1), 26862.

[14] Zhao, P., Liu, X., Su, X., Wu, D., Li, Z., Kang, K., ... & Zhu, A. (2025). Probabilistic Contingent Planning Based on Hierarchical Task Network for High-Quality Plans. Algorithms, 18(4), 214.

[15] Zheng, Z., Liu, K., & Zhu, X. (2025). Machine learning-based prediction of metal-organic framework materials: A comparative analysis of multiple models. arXiv preprint arXiv:2507.04493.

[16] Feng, C., Bačić, B., & Li, W. (2024, December). Sca-lstm: A deep learning approach to golf swing analysis and performance enhancement. In International Conference on Neural Information Processing (pp. 72-86). Singapore: Springer Nature Singapore.

[17] Ning, Z., Zeng, H., & Tian, Z. (2025). Research on data-driven energy efficiency optimisation algorithm for air compressors. In Third International Conference on Advanced Materials and Equipment Manufacturing (AMEM 2024) (Vol. 13691, pp. 1068–1075). SPIE.

[18] Li, X., & Fang, L. (2025). The Impact of Culturally Adaptive Feedback in Information-Interaction Systems on Language-Learning Motivation and Outcomes. Journal of Educational Theory, 2(1), 1–7.

[19] Siye Wu, Lei Fu, Runmian Chang, et al. Warehouse Robot Task Scheduling Based on Reinforcement Learning to Maximize Operational Efficiency. TechRxiv. April 18, 2025. DOI: 10.36227/techrxiv.174495431.17315991/v1

[20] Wang, J., Zhang, Z., He, Y., Song, Y., Shi, T., Li, Y., ... & He, L. (2024). Enhancing Code LLMs with Reinforcement Learning in Code Generation. arXiv preprint arXiv:2412.20367.

[21] Zhou, Y., Shen, J., & Cheng, Y. (2025). Weak to strong generalization for large language models with multi-capabilities. In The Thirteenth International Conference on Learning Representations.

[22] Bačić, B., Vasile, C., Feng, C., & Ciucă, M. G. (2024). Towards nation-wide analytical healthcare infrastructures: A privacy-preserving augmented knee rehabilitation case study. arXiv preprint arXiv:2412.20733.

[23] Wang, H., Tang, H., Leng, N., & Yu, Z. (2025). A Machine Learning-Based Study on the Synergistic Optimization of Supply Chain Management and Financial Supply Chains from an Economic Perspective. arXiv preprint arXiv:2509.03673.